\documentclass[lettersize,journal]{IEEEtran}
\usepackage{amsmath,amsfonts}
\usepackage{algorithmic}
\usepackage{array}
\usepackage{textcomp}
\usepackage{stfloats}
\usepackage{url}
\usepackage{verbatim}
\usepackage{graphicx}
\def\BibTeX{{\rm B\kern-.05em{\sc i\kern-.025em b}\kern-.08em
    T\kern-.1667em\lower.7ex\hbox{E}\kern-.125emX}}
\usepackage{balance}

\usepackage{wrapfig}
\usepackage{amssymb}
\usepackage{nicefrac}       % compact symbols for 1/2, etc.
\usepackage{subfigure}
\usepackage{booktabs}
\usepackage{bm}
\usepackage{multirow}
\usepackage[linesnumbered,ruled]{algorithm2e}
\usepackage{orcidlink}
\usepackage{xcolor} % 加载颜色包

\definecolor{roy}{rgb}{0.0, 0.0, 0.0}
\definecolor{howard}{rgb}{0.0, 0.0, 0.0}

\definecolor{ccr}{RGB}{0,0,255}  
\usepackage{hyperref}
\hypersetup{hypertex=true,
	colorlinks=true,
	linkcolor=ccr,
	urlcolor=black,
	citecolor=ccr}

\begin{document}
\title{Enhancing Perception Quality in Remote Sensing Image Compression via Invertible Neural Network}
\author{Junhui Li${^{\orcidlink{0000-0001-9143-1321}}}$, Xingsong Hou${^{\orcidlink{0000-0002-6082-0815}}}$
	\thanks{Manuscript received xx, 2024; This work was supported by the National Natural Science Foundation of China under Grant 62272376.  \textit{(Corresponding authors: Xingsong Hou.)}		
		
			The authors are with the School of Information and Communications Engineering, Xi’an Jiaotong University, Xi’an 710049, China (e-mail: mlkkljh@stu.xjtu.edu.cn; houxs@mail.xjtu.edu.cn).	
			
%			Wenke Sun is with CRRC Qingdao Sifang Rolling Stock Research Institute Co., Ltd., Qingdao, 266031, China.
}}

\markboth{Journal of \LaTeX\ Class Files,~Vol.~18, No.~9, August~2024}%
{How to Use the IEEEtran \LaTeX \ Templates}

\maketitle

\begin{abstract}
Despite the impressive perceptual or fidelity performance of existing image compression algorithms, they typically struggle to strike a balance between perceptual quality and high image fidelity. To address this challenge, we propose a novel invertible neural network-based remote sensing image compression (INN-RSIC) method. Our approach captures the compression distortion of an existing image compression algorithm and encodes it as a set of Gaussian-distributed latent variables via an invertible neural network (INN). This ensures that the compression distortion in the decoded image remains independent of the ground truth. By leveraging the inverse mapping of the INN, we can input the decoded image along with a set of randomly resampled Gaussian-distributed variables into the inverse network, effectively generating enhanced images with improved perceptual quality.
To accurately learn and represent compression distortion, we utilize channel expansion, Haar transformation, and invertible blocks in constructing the INN. Additionally, we introduce a quantization module (QM) to mitigate the impact of format conversion, thereby enhancing the framework's generalization capabilities and improving the perceptual quality of the enhanced images. Extensive experiments demonstrate that the proposed INN-RSIC achieves a superior balance of perceptual quality and high image fidelity compared to existing image compression algorithms. Moreover, the developed INN-based enhancer operates as a plug-and-play (PnP) method, allowing seamless integration with existing image compression algorithms to improve their decoding perceptual quality. Our code and pre-trained models are available at \url{xxxx}.	
\end{abstract}

\begin{IEEEkeywords}
	Remote sensing image compression, invertible neural network, Haar transformation, image enhancement.
\end{IEEEkeywords}

\section{Introduction}
\IEEEPARstart{R}{emote} sensing (RS) images have been widely applied in various fields, including urban planning \cite{10014660}, biosphere monitoring \cite{9325523}, and land resource management \cite{10287366}. However, RS images are typically characterized by high content, high resolution, and large size \cite{xiang2023remote, li2021image}. Furthermore, with the advancement of sensor technology and the enhancement of the image acquisition capability of satellite and airborne equipment,  the necessity of storing or transmitting RS images is also growing in importance. To cope with these challenges, high-resolution RS images are usually compressed before they are stored and transmitted to the ground. In this case, it is critical to achieve low-bitrate compression while obtaining the decoded images with high perceptual quality.

%(遥感数据很多，很多学者开始研究， 尤其在军事应用中， 对关键目标保真尤为重要)
Traditional image compression algorithms, including JPEG2000 \cite{taubman2002jpeg2000}, BPG \cite{bpg2017}, and  VVC \cite{VVC2021}, have been instrumental in facilitating the storage and transmission of image data. However, image compression with these algorithms inevitably suffered from undesired blocking, ringing artifacts, and blurring \cite{pan2023hybrid, cheng2020learned}, which may highly influence the perception quality.

With the development of deep learning techniques, learning-based image compression methods have made significant progress in obtaining high rate-distortion (RD) performance \cite{zhang2023global, pan2023hybrid, 8793171, tang2022joint, 9088290, 8931632, liu2023learned, jiang2023mlic}. Notably, in natural image compression,  Ballé \textit{et al.} \cite{47602} proposed to view additional side information as a hyper-prior entropy model to estimate a zero-mean Gaussian distribution, laying the foundation for subsequent improvements in entropy modeling. Cheng \textit{et al.} \cite{cheng2020learned} further introduced discretized Gaussian mixture likelihoods to realize the distribution estimation of the entropy model, resulting in impressive decoding performance. Meanwhile, He \textit{et al.} \cite{he2022elic} developed a spatial-channel contextual adaptive model, enhancing compression performance without sacrificing computational speed. Liu \textit{et al.} \cite{liu2023learned} leveraged the local modeling ability of convolutional neural networks (CNNs) and the non-local modeling strengths of Transformers to develop the encoder and decoder networks. A channel-squeezing-based entropy model was further proposed to enhance RD performance. Jiang \textit{et al.} \cite{jiang2023mlic} proposed capturing the channel-wise, local spatial, and global spatial correlations present in latent representations to develop a comprehensive entropy model, which is employed to design the competitive image compression method MLIC$^+$.
Despite the success of learning-based compression algorithms in natural scenes, RS images present unique challenges due to their rich texture, weak correlation, and low redundancy compared to natural images \cite{lu2017exploring, han2023edge}. Preserving visually pleasing decoded images while achieving high compression bitrates remains a significant challenge in RS image compression. \par

To tackle the challenge, Zhang \textit{et al.} \cite{zhang2023global} introduced a multi-scale attention module to enhance the network's feature extraction capability, and developed an improved entropy model by using global priors and anchored-stripe attention. 
Pan \textit{et al.} \cite{pan2023coupled} resorted to generative adversarial networks (GANs) to independently reconstruct the image content and detailed textures, subsequently fusing these features to achieve low-bitrate RS image compression.  Additionally, Xiang \textit{et al.} \cite{10379598} utilized discrete wavelet transform to separate image features into high- and low-frequency components, and designed compression networks to enhance the model's representation ability of both types of features. \par
\begin{figure*} %{r}{0.6\textwidth} % {r} 表示将图片放在右侧，{0.5\textwidth} 表示图片宽度为正文宽度的一半
	\centering
	\includegraphics[scale=1.1]{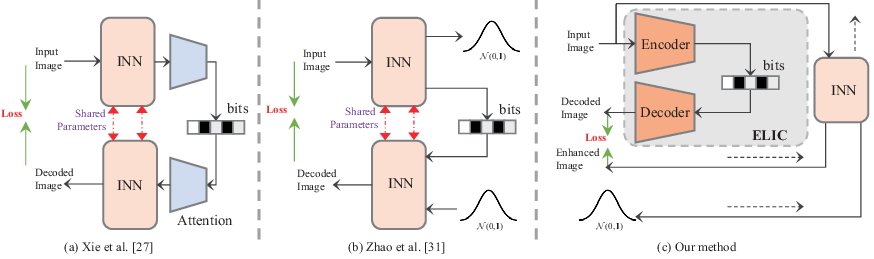} % 修改为你的图片文件名
	\caption{Overview of existing INN-based image compression methods. For simplicity, entropy models are omitted. It can be seen that our method is a plug-and-play (PnP) approach that does not require retraining of existing compression models. Our method can obtain both decoded images with high PSNR and MS-SSIM output from the original compression algorithm (\textit{e.g.}, ELIC) and enhanced images with high perceptual quality.
		}
	\label{Fig: INN_com}
\end{figure*}
\begin{figure} %{r}{0.6\textwidth} % {r} 表示将图片放在右侧，{0.5\textwidth} 表示图片宽度为正文宽度的一半
	\centering
	\includegraphics[scale=0.9]{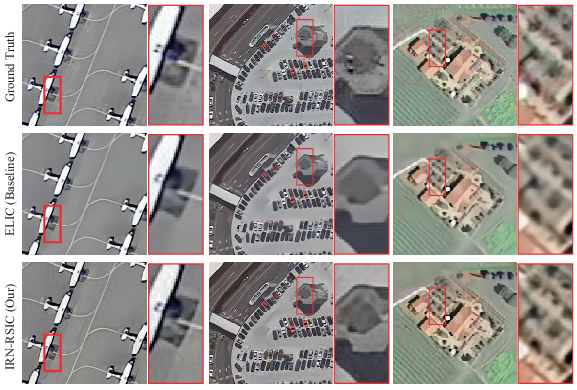} % 修改为你的图片文件名
	\caption{Visualization of the decoded images by ELIC \cite{he2022elic} and the proposed INN-RSIC.}
	\label{Fig:com_base_UC}
\end{figure}

%Typically, when developing a learning-based image compression approach, two critical factors are considered. Firstly, a reduction in the correlation among latent representation coefficients achieved by the encoder network enhances the potential for conserving bitrate during entropy coding. Secondly, an accurate estimation of the probability distribution of these coefficients using an entropy model allows for more efficient utilization of the bitstream, providing enhanced control over the required bitrate for encoding the latent representations \cite{hu2021learning}. \par
In summary, those above deep learning-based image compression methods typically achieve image compression through CNN blocks \cite{cheng2020learned, he2022elic, 47602, tang2022joint, zhang2023global, 10379598} or Transformer blocks \cite{zou2022devil, qian2022entroformer, 10120984, xie2021enhanced, 10516594}, followed by optimization using distance measurements like L$ _1 $, L$ _2 $, or adversarial loss \cite{mentzer2020high}. Despite their impressive performance, these methods usually struggle to generate satisfactory perception quality while maintaining high image fidelity. Recently, invertible neural networks (INN) have emerged as a new paradigm for image generation, demonstrating remarkable performance across various applications \cite{9467808, 10256132, 9298920, 9662053}. Building on this success, several approaches have integrated INN into image compression to capture richer texture details \cite{xie2021enhanced,10256132}. Fig. \ref{Fig: INN_com} illustrates the overview of existing INN-based compression methods. For instance, in  \cite{xie2021enhanced}, an enhanced INN-based encoding network was devised to improve the decoding performance of natural image compression. In \cite{10256132}, the authors utilized INN to develop an invertible image generation module, aiming to prevent information loss and propose a competitive low-bit-rate compression algorithm. Unlike these methods,  the proposed approach focuses on leveraging the invertible capabilities of INN to recover more texture information of images decoded by an existing compression algorithm, without requiring additional bitstreams and model retraining. It is expected to sample some texture details from Gaussian distributed variables conditioned on the decoded image. Fig. \ref{Fig:com_base_UC} illustrates the perception comparison between the decoded images and the enhanced images. One can see that the proposed INN-RSIC exhibits a high capability in recovering texture-rich details without additional bitrates. \par
%\begin{figure}[!htbp]
%	\centering
%	\subfigure[The pipline of the existing image compression method]{
	% \begin{minipage}[t]{3.2 in}
		% \centering
		% \includegraphics[scale=1]{./images/tradition_method.png}
		% %\caption{fig1}
		% \end{minipage}%
	%	}%	
%
%	\subfigure[The pipline of proposed method]{
	% \begin{minipage}[t]{3.2in}
		% \centering
		% \includegraphics[scale=1]{./images/Our_method.png}
		% %\caption{fig2}
		% \end{minipage}
	%	}%
%	\centering
%	\caption{The pipeline of the existing image compression method (a) and the proposed INN-RSIC (b). The proposed method focuses on recovering high-quality decoded images without additional bitrates. }
%	\label{Fig:tradition_our}
%\end{figure}

In this paper, we propose the invertible neural network-based remote sensing image compression (INN-RSIC). Specifically, we utilize an invertible forward network with a conditional generation module (CGM) to encode compression loss information of an existing image compression algorithm into Gaussian-distributed latent variables. As a result, it is expected to sample some of the lost prior information from Gaussian distribution, thereby facilitating the recovery of visually enhanced images during the inverse mapping. Additionally, to effectively learn compression distortion, we adopt channel expansion and Haar transformation \cite{haar1909theorie} to separate high and low frequency and introduce a quantization module (QM) to reduce the impact of the reconstruction quality due to data type conversion in the inference stage. \par
The primary contributions of this paper can be summarized as follows: 
\begin{itemize}
	\item[$\bullet$] The proposed INN-RSIC is the first attempt to model image compression distortion of an image compression method using invertible transforms. It serves as a PnP method to obtain highly perceptible decoded images while preserving the performance of the baseline algorithm.
	
	\item[$\bullet$] We develop a novel, effective yet simple architecture with channel expansion, Haar transformation, and invertible blocks. This architecture enables projecting compression distortion into a case-agnostic distribution, so that compression distortion information can be obtained based on samples from a Gaussian distribution.
	
	\item[$\bullet$] CGM is introduced to split and encode the compression distortion information from ground truth into the latent variables conditioned on the synthetic image. This process generates pattern-free synthetic images and enhances the texture details of the reconstructed images during the inverse mapping.
	
	\item[$\bullet$] Extensive experiments indicate that the proposed INN-RSIC achieves a superior balance of perceptual quality and high image fidelity compared to existing state-of-the-art image compression algorithms. Our method offers a novel perspective for image compression algorithms to improve their perception quality.  \par
\end{itemize}
%The remainder of the paper is organized as follows. Section \ref{Sec:related_work} provides a comprehensive review of the related work. Section \ref{Sec:Proposed} describes the proposed INN-RSIC in detail. Section \ref{Sec:Experiments} presents the experimental results and analysis and Section \ref{Sec:Conclusion} concludes the paper. \par
\section{Related work} \label{Sec:related_work}
%In this section, we introduce related work in two aspects. The first is end-to-end RS image compression, which has emerged as a prominent field of interest in recent years. The second is INNs, which have provided fundamental inspiration for our research work.
\subsection{End-to-End RS Image Compression}
In RS image compression, many researchers have directed their attention towards network structure and image transformation techniques. For instance, Pan \textit{et al.} \cite{pan2023coupled} adopted GANs to independently decode the image content and detailed textures, later combining these features to achieve low-bitrate compression of RS images. More recently, Xiang \textit{et al.} \cite{10379598} leveraged the discrete wavelet transform to separate image features into high- and low-frequency components. They then designed compression networks to bolster the model's ability to present these features at different frequencies to recover texture-rich details.

%% 提高熵模型的性能
Moreover, in the pursuit of advancing network modules for decoding more texture-rich details of RS image compression, some studies aim to improve the estimated accuracy of the entropy model. Pan \textit{et al.} \cite{pan2023hybrid} proposed a hybrid attention network to improve the prediction accuracy of the entropy model and enhance the representation capabilities of both the encoder and decoder. In \cite{xiang2023remote}, a long-range convolutional network was developed as a network model to improve the decoding performance of RS image compression. Fu \textit{et al.} \cite{rs15082211} considered the local and non-local redundancies presented in RS images and developed a mixed hyperprior network to explore both, thus improving the accuracy of entropy estimation.
%Besides,  Zhang \textit{et al.} \cite{zhang2023global} introduced a multi-scale attention module to enhance the network's feature extraction capability, and employed global priors and anchored-stripe attention to improve the entropy model. 

In summary, the above RS image compression algorithms mainly focus on designing complex networks for compact latent variable representation and more accurate entropy estimation to achieve texture-rich decoding results. Different from the above methods, we innovatively project the compression distortion loss of decoded images from an image compression method into Gaussian-distributed latent variables in the forward network, while combining the Gaussian distribution with the decoded images in the inverse network to obtain texture-rich enhanced images.

\subsection{Invertible Neural Network (INN)} INN offers several advantages, including bijective mapping, efficient mapping access, and computable Jacobi matrices, making them promising for various machine learning tasks \cite{NEURIPS2020_1cfa81af, 10227600, huang2022winnet, LIU2023109822, 9858176, 10445438, NEURIPS2019_70a32110}. 
For instance, Zhao \textit{et al.} \cite{9467808} employed INN to generate invertible grayscale images by separating color information from color images and encoding it into Gaussian-distributed latent variables. This approach ensures that the color information lost during grayscale generation remains independent of the input image. Similarly, Xiao \textit{et al.} \cite{Inverse_rescale} developed invertible models to generate valid degraded images while transforming the distribution of lost contents to a fixed distribution of a latent variable during forward degradation. The restoration was then made tractable by applying the inverse transformation on the generated degraded image along with a randomly drawn latent variable. Liu \textit{et al.} \cite{LIU2023109822} considered image degradation and super-resolution as a pair of inverse tasks and replaced the generators in GAN with INN for unpaired image super-resolution. In addition, INN technology has been leveraged to improve the transformation between the image space and the latent feature space, thereby mitigating information loss \cite{xie2021enhanced}. For instance,  Xie \textit{et al.} \cite{xie2021enhanced} proposed an enhanced INN-based encoding network for better image compression. Gao \textit{et al.} \cite{10256132} integrated INN into the development of an invertible image generation module to prevent information loss and developed a competitive low-bitrate image compression algorithm. 

\section{The Proposed INN-RSIC} \label{Sec:Proposed}
Compared to natural images, RS images usually contain richer textures, which makes acquiring RS images at low bit rates more challenging \cite{han2023edge}. Therefore, the proposed INN-RSIC aims to utilize INN to encode the lost compression distortion of an existing image compression algorithm into a set of latent variables $\mathbf{z}$, following a pre-defined distribution, such as a Gaussian distribution. As a result, the distribution of latent variables becomes independent of the distribution of the input image. In the enhanced reconstruction stage of our framework (\textit{i.e.}, the inverse mapping of the proposed INN-RSIC), a new set of randomly sampled latent variables $\vec{\mathbf{z}}$ can effectively represent the lost compression distortion information to some extent. Thus, we can simply feed the re-sampled $\vec{\mathbf{z}}$, along with the decoded images, to the INN-RSIC of the inverse network for enhanced images.

Generally, the distribution of compression distortion information varies among different learning-based methods, as each exhibits distinct preferences in learning data distribution. For instance, some algorithms excel in achieving high performance on texture-rich images, while others have the opposite. Therefore, we focus here on the compression distortion distribution of the impressive image compression algorithm ELIC at different bitrates. Unfortunately, pinpointing the distortion distribution directly is challenging. Therefore, we choose to capture the distortion by investigating the relationship between the ground truth and the decoded image. In this way, we can indirectly explore the distortion distribution inherent in the compression algorithm.\par

\begin{figure*}[!htbp]
	\centering
	\includegraphics[scale=1.12]{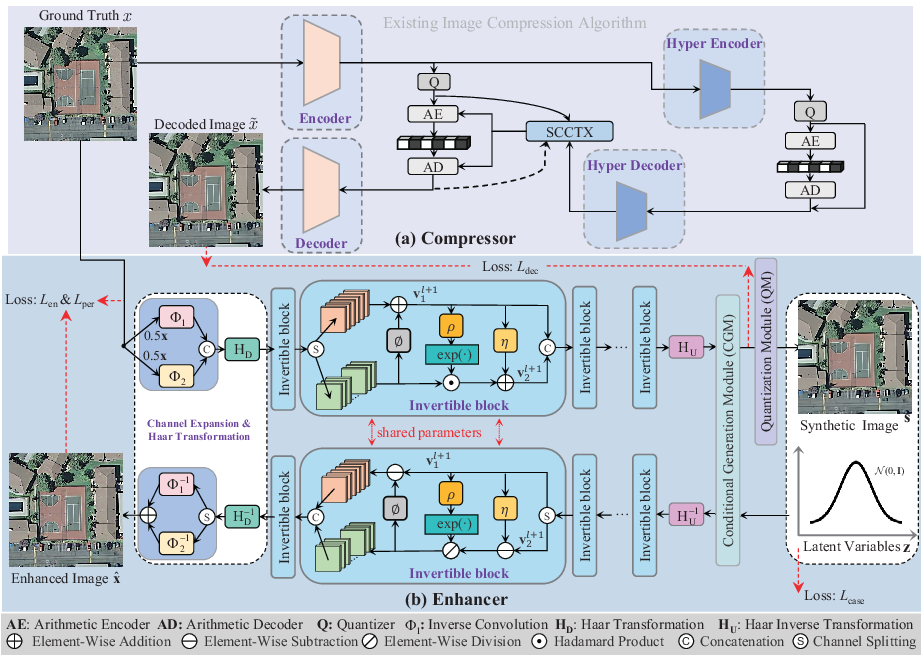}
	\caption{The architecture of the proposed INN-RSIC. Compressor refers to the competitive image compression algorithm \cite{he2022elic}.}
	\label{Fig:main_framework}
\end{figure*}

Specifically, the architecture of the proposed INN-RSIC is depicted in Fig. \ref{Fig:main_framework}, consisting of a compressor and an enhancer. Given an input image $\mathbf{x}$, we utilize the competitive image compression algorithm ELIC \cite{he2022elic} as the compressor to generate the decoded image $\tilde{\mathbf{x}}$, which serves as a label to constrain the generated synthetic image $\mathbf{s}$ for training the enhancer. The enhancer mainly comprises two streams: the forward and inverse networks, denoted as $f_{\mathbf{\Theta}}(\cdot)$ and $f_{\mathbf{\Theta}}^{-1}(\cdot)$, respectively, where $\mathbf{\Theta}$ represents the network parameters. \par
\subsection{INN Architecture}
\subsubsection{Invertible Block}
The invertible block layer is a crucial component in invertible architectures, acting as a bridge to adapt between two different distributions via its trainable parameters. Fig.  \ref{Fig:main_framework}(b) depicts the structure of an invertible block. An inverse block operates on an input feature map \(\mathbf{v} \in \mathbb{R}^{H \times W \times C}\), where \(H\), \(W\), and \(C\) denote height, width, and the number of channels, respectively. In each invertible block, \(\mathbf{v}\) is splitted along the channel dimension into two parts \([\mathbf{v}_1, \mathbf{v}_2]\), where \(\mathbf{v}_1 \in \mathbb{R}^{H \times W \times C_1}\) and \(\mathbf{v}_2 \in \mathbb{R}^{H \times W \times C_2}\), and \(C_1 + C_2 = C\). 
To obtain the input features of the $(l+1)$-th invertible block, inverse transformations are applied to both segments using learnable scale and shift parameters by
\begin{equation}
\begin{aligned} 
	\mathbf{v}_{1}^{l+1} &= \mathbf{v}_1^{l} + \phi(\mathbf{v}_2^{l}),  \\
	\mathbf{v}_2^{l+1} &= \mathbf{v}_2^{l} \odot \exp(\rho(\mathbf{v}_1^{l+1})) + \eta(\mathbf{v}_1^{l+1}),
\end{aligned}
\label{Eq:Inverse_block}
\end{equation}
where $ \odot$ refers to Hadamard product. The $\varphi(\cdot)$, $\rho(\cdot)$, and $\eta(\cdot)$ are achieved by the dense blocks \cite{wang2018esrgan}. Therefore, the inverse  transformation can be derived from Eq. (\ref{Eq:Inverse_block}) by 
\begin{equation}
	\begin{aligned} 
		\mathbf{v}_1^{l} &= \mathbf{v}_1^{l+1} - \phi(\mathbf{v}_2^{l}), \\
	\mathbf{v}_2^{l} &= (\mathbf{v}_2^{l+1} - \eta(\mathbf{v}_1^{l+1})) \oslash \exp(\rho(\mathbf{v}_1^{l+1})), \\
	\end{aligned} 
\end{equation}
where $\oslash$ denotes element-wise division.\par

Besides, as illustrated in Fig. \ref{Fig:main_framework}(b), the forward network outputs the synthetic image $\mathbf{s}$ and the latent variable $\mathbf{z}$. Following the studies \cite{10256132, xie2021enhanced}, we assume that $\mathbf{z}$ follows a Gaussian distribution, which can be randomly sampled from the same distribution for the inverse mapping. Consequently, in the training stage, the synthetic image $\mathbf{s}$ and $\mathbf{z}$ can be merged and fed into the inverse network for model optimization. In the inference stage, the decoded image $\tilde {\mathbf{x}}$ and resampled $\vec{\mathbf{z}} \sim \mathcal{N}(0, \mathbf{I})$ can be combined and fed into the inverse network for image enhancement.

\subsubsection{Channel Expansion and Haar Transformation}
Within each invertible block, the channels of both its input and output signals have the same number. However, in addition to generating the synthetic image, we also need to produce an expanded output to derive the Gaussian latent variable $\mathbf{z}$. To achieve this, as shown in Fig. \ref{Fig:main_framework}(b), we double the number of channels of the input image by using $1\times1$ convolutional layers.  In addition, Haar transformation \cite{haar1909theorie} is used here to separate the compression distortion information by splitting the high and low frequencies, which are then fed into the invertible blocks.

Specifically,  in the forward network, given the ground truth $\mathbf{x}$, the feature $\mathbf{F}$ fed into the first invertible block can be expressed as
\begin{equation}
	\mathbf{F} = H_D(\operatorname{Concat}(\Phi_1(\mathbf{x}/2.0), \Phi_2(\mathbf{x}/2.0))), \label{Eq:channel_exp}
\end{equation}
where $H_D(\cdot)$ denotes Haar function,  and $\operatorname{Concat}(\cdot)$ refers to a concatenation operation. $\Phi_1(\cdot)$ and $\Phi_2(\cdot)$ refer to two $1\times1$ convolution layers. After the last invertible block, the inverse Haar function $H_U(\cdot)$
is adopted to transform the features from the frequency domain to the spatial domain.

As a result, at the end of the forward network, we can thus obtain six-channel signals with the same size as the input image, and with CGM, we derive both the synthetic image $\mathbf{\Theta}$ and Gaussian latent variables $\mathbf{z}$, which are then combined and fed into the inverse network for image recovery. 

In the inverse network, the inverse processing of Eq. (\ref{Eq:channel_exp}) can be formulated as
\begin{equation}
	\begin{aligned}
		\mathbf{p}_i &= \operatorname{Split}(H_D^{-1}(\mathbf{F})), ~~ i \in \{1,2\}, \\
		\mathbf{x} &= \Phi^{-1}_1(\mathbf{p}_1) + \Phi^{-1}_2(\mathbf{p}_2),  
	\end{aligned}	
\end{equation}
where  $\Phi^{-1}_1(\cdot)$, $\Phi^{-1}_2(\cdot)$, and $H_D^{-1}(\cdot)$ refer to the inverse functions of  $\Phi_1(\cdot)$, $\Phi_2(\cdot)$, and $H_D(\cdot)$, respectively. Next, we will illustrate CGM in detail.

%To enhance the encoding of compression distortion information from input images into Gaussian-distributed latent variables, predicated on their grayscale equivalents, the encoding scheme integrates a conditional splitting layer at its culmination. This inclusion aims to augment the color reconstruction mechanism, enabling it to adjust according to the peculiarities of the input image. Initiated with the tri-channel tensor emanating from the terminal invertible block—dimensionally analogous to the input image—this process partitions \(y\) into a duo of segments: a singular-channel grayscale observation \(g\) and a bi-channel latent representation \(z\) embodying the color data. This procedure necessitates a revision of the previously delineated affine coupling layer, to instigate a correlation between \(g\) and \(z\). The architecture of this conditional splitting layer is delineated in Fig. 4. During the forward traversal, \(y\) segregates into the latent entities \(\tilde{z}\) and the grayscale \(g\). A unidirectional affine coupling layer is employed to transmute \(\tilde{z}\) into standardized Gaussian-distributed variables \(z\), in the manner delineated as follows:

\subsubsection{Conditional Generation Module (CGM)}
\begin{figure} %{r}{0.5\textwidth} % {r} 表示将图片放在右侧，{0.5\textwidth} 表示图片宽度为正文宽度的一半
	\centering
	\includegraphics[scale=1.1]{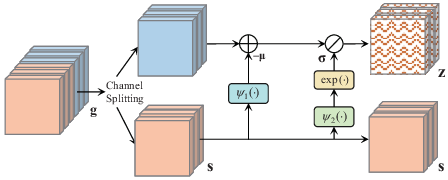} % 修改为你的图片文件名
	\caption{Illustration of the conditional generation module (CGM).}
	\label{Fig:Condition_layer}
\end{figure}

As illustrated above, we aim to encode the compression distortion information of decoded RS images into a set of Gaussian-distributed latent variables, where the mean and variance are conditioned on the synthetic image. This conditioning enables the reconstruction process to be image-adaptive. To achieve this, as illustrated in Fig. \ref{Fig:main_framework}(b), we introduce  CGM at the end of the forward network. Specifically,  to establish the dependency between \( \mathbf{s} \) and \( \mathbf{z} \), as depicted in Fig. \ref{Fig:Condition_layer}, the output six-channel tensor \( \mathbf{g} \)
from the last invertible block is further divided by CGM into two parts: a three-channel synthetic image \( \mathbf{s} \) and a three-channel latent variable \( \tilde{\mathbf{z}} \) representing the compression distortion information.

In the forward mapping, inspired by \cite{9467808}, we normalize \( \tilde{\mathbf{z}} \) into standard Gaussian-distributed variables \( \mathbf{z} \) by $\mathbf{z} = (\tilde{\mathbf{z}} - \bm{\mu}) \oslash \bm{\sigma}$, where $\mathbf{z} \sim \mathcal{N}(0, \mathbf{I})$, where the mean and scale of $\mathbf{s}$ can be calculated by
\begin{equation}
\begin{aligned}
	\boldsymbol{\mu} &= \psi_1(\mathbf{s}), \\
	\boldsymbol{\sigma} &= \exp(\psi_2(\mathbf{s})) ,
\end{aligned}
\end{equation}
where \( \psi_1(\cdot) \)  and \( \psi_2(\cdot) \) are achieved by the dense block \cite{wang2018esrgan}.  Hence, its reverse mapping can be formulated as $\tilde{\mathbf{z}} = \mathbf{z} \odot \bm{\sigma} + \bm{\mu}$, where $\tilde{\mathbf{z}} \sim \mathcal{N}(\bm{\mu}, \bm{\sigma})$. In this way, we encode the compression distortion information into the latent variables, whose distribution is conditioned on the synthetic image. The inverse network is similar to \cite{Inverse_rescale,9467808}, where we sample a set of random variables from  Gaussian distribution conditioned on the synthetic images to reconstruct the enhanced images.

\subsubsection{Quantization Module (QM)}
To ensure compatibility with common decoded image storage formats such as RGB (8 bits for each R, G, and B color channels), as shown in Fig. \ref{Fig:main_framework}(b), we integrate QM after CGM. This module converts the floating-point values of the produced synthetic images into 8-bit unsigned integers by a rounding operation for quantization. However, it's essential to acknowledge a significant obstacle: the quantization module is inherently non-differentiable. To cope with the challenge, we employ the pass-through estimator technique used in \cite{he2022elic} to ensure that INN-RSIC is efficiently optimized during the training process. Subsequently, in the inference stage, the decoded image can be reasonably fed into the inverse network for image enhancement.

\subsection{Optimization Strategy} \label{Sec:Opt_strategy}
\subsubsection{Compression Optimization Loss}
As we aim to capture the compression distortion of ELIC \cite{he2022elic}, here we provide a brief overview of its loss function. Specifically, to balance the compression ratio and the image quality of the decoded image of ELIC, the loss function can be formulated as 
\begin{equation}
	\mathcal{L}_1 = R + \lambda D( \mathbf{x}, \tilde {\mathbf{x}} ) ,
	\label{Eq:loss_mse}
\end{equation}
where the rate $R$ refers to the entropy of the quantized latent variables of ELIC. Meanwhile, $D(\cdot, \cdot)$ denotes the similarity between the input image and the decoded image, typically measured by using mean squared error (MSE).  Different compression rates can be achieved by adjusting the hyperparameter $\lambda$, where the higher the value of $\lambda$, the higher the bits per pixel (bpp) and the better the image quality.

\subsubsection{Forward Presentation Loss} The forward network is primarily focused on enabling the proposed model to capture the compression distortion distribution of the decoded image. This is achieved by establishing a correspondence between the input image \(\mathbf{x}\) and the decoded image \(\tilde {\mathbf{x}}\), as well as a case-agnostic distribution \(p(\mathbf{z})\) of \(\mathbf{z}\). It can be realized by developing two loss functions: decoded image loss and case-agnostic distribution loss. 

\subsubsection*{\textbf{Decoded image loss}} To generate the guidance images for training the forward network, ELIC is adopted to generate the labeled image $\tilde {\mathbf{x}}$, which can be formulated as
\begin{equation}
	\tilde {\mathbf{x}} = \operatorname{ELIC}(\mathbf{x}, \lambda). \label{Eq:ELIC}
\end{equation}
Thereafter, to make our model follow the guidance, we drive the synthetic image $\mathbf{s} = f_{\mathbf{\Theta}}(\mathbf{x})$ to resemble $\tilde {\mathbf{x}}$, which can be derived by
\begin{equation}
	L_{\text{dec}} = ||f_{\mathbf{\Theta}}(\mathbf{x})-\tilde {\mathbf{x}}||^2.
\end{equation}

\subsubsection*{\textbf{Case-agnostic distribution loss}} To regularize the distribution of \(\mathbf{z}\), we maximize the log-likelihood of $p(\mathbf{z})$, inspired by \cite{9467808}. Thus, the loss function to constrain the latent variables $\mathbf{z}$ can be formulated as 
\begin{equation} %\small
	\begin{aligned}
		L_{\text{case}} &= -\log( p(\mathbf{z})) \\
	&= -\log\left(\frac{1}{(2\pi)^{M/2}} \exp\left(-\frac{1}{2} ||\mathbf{z}||^2\right)\right),
	\end{aligned}
\end{equation}
where $M$ is the dimensionality of $\mathbf{z}$. This loss function penalizes the normalized latent variables $\mathbf{z}$ to follow standard Gaussian distribution. Consequently, in the inverse processing, we can randomly sample a set of Gaussian-distributed variables along with the synthetic image to derive the enhanced image.

\subsubsection{Reverse reconstruction loss}
The inverse network aims to guide the model in recovering visually appealing images using randomly sampled latent variables and synthetic images. This is achieved by developing two loss functions: enhancement reconstruction loss and quality perception loss.	

\subsubsection*{\textbf{Enhancing reconstruction loss}} In theory, the synthetic image can be perfectly restored to the corresponding ground truth version through the inverse network of INN because there is no information omission. However, in practice, the decoded image is not generated by the forward network of the proposed INN-RSIC but with an image compression method. That is, the synthetic image should be stored in 8-bit unsigned integer format so that the decoded image can be used directly instead of the synthetic image here for image detail enhancement during the inference stage.
To address this, we adopt QM at the end of the forward network during training. Consequently, by penalizing the discrepancy between the reconstructed image and the ground truth, we can derive the enhancing reconstruction loss given as

\begin{equation}
	L_{\text{en}} = || f_{\mathbf{\Theta}}^{-1}(\mathbf{s}, \vec{\mathbf{z}}) - \mathbf{x} ||^2, \label{Eq: loss_en}
\end{equation}
where $\vec{\mathbf{z}}$ denotes the re-sampled latent variables from standard Gaussian distribution, and the synthetic image can be derived by $\mathbf{s}= \operatorname{QM}(f_{\mathbf{\Theta}}(\mathbf{x}))$.

\subsubsection*{\textbf{Quality perception loss}} To improve the perceptual performance of the network by estimating distances in predefined feature space rather than image space, a perceptual loss function called $L_{\text{per}}$ is used here. In other words, the feature space distance is represented by an optimization function, which serves as a driver for the network to perform image reconstruction while retaining a feature representation similar to the ground truth.
Concretely, the learning perceptual image patch similarity (LPIPS) \cite{zhang2018unreasonable} is utilized to indicate the perceptual loss, which is defined as
\begin{equation}
	L_{\text{per}} = ||\operatorname{Vgg}(\hat{\mathbf{x}}) - \operatorname{Vgg}(\mathbf{x})||^2 , \label{Eq: vgg_loss}
\end{equation}
where \( \operatorname{Vgg}(\cdot) \) denotes a feature function that leverages the features extracted at the ``Conv4-4" layer of VGG19 to
penalize the contrast similarity between $\hat{\mathbf{x}}$ and $\mathbf{x}$, as adopted in \cite{10.1145/3272127.3275080}.

%that maps the image space to the feature space by leveraging the pre-trained VGG-19 network \cite{simonyan2014very}. 

Therefore, the total loss function can be given by
\begin{equation}
	L_{\text{total}} = \lambda_1 L_{\text{dec}} + \lambda_2 L_{\text{case}} + \lambda_3 L_{\text{en}} + \lambda_4 L_{\text{per}},
\end{equation}
where $\lambda_1, \lambda_2, \lambda_3$, and $\lambda_4$ are hyperparameters used to balance the different loss terms.

Thus, after obtaining the trained model, in the inference stage, given a decoded image $\tilde {\mathbf{x}}$ by using Eq. (\ref{Eq:ELIC}), we can derive its enhanced image $\hat {\mathbf{x}}$ by $\hat {\mathbf{x}} = f_{\mathbf{\Theta}}^{-1}(\tilde {\mathbf{x}}, \vec{\mathbf{z}})$, where $\vec{\mathbf{z}}$ refers to the latent variables sampled from Gaussian distribution $ \vec {\mathbf{z}} \sim \mathcal{N}(0, \mathbf{I})$. It can be observed that there is no additional bitrates acquirement for the enhancement processing of the decoded image. The training and inference stages of the proposed INN-RSIC are summarized in Algorithm \ref{alg:A}. 
\begin{algorithm} [!htbp] 
	\caption{Processing of INN-RSIC}  	
	\label{alg:A}  	
	\hspace*{0.02in}{\bf Training Stage:}\\
	\hspace*{0.2in}\small{Input:}
	$ \mathbf{x}, \lambda $ \\
	\hspace*{0.2in}\small{Output:} 
	$\tilde {\mathbf{x}}, \mathbf{\Theta}$ 	
	\begin{algorithmic}[1] 
			\STATE  {\small {\textbf{Procedure:}  Compressor}$ (\mathbf{x}, \lambda)$}
			\begin{ALC@g}		
					\STATE \small{$\tilde {\mathbf{x}} = \operatorname{ELIC}(\mathbf{x}, \lambda), $}
				\end{ALC@g}	
			\STATE {\small {\textbf{Return}}: $\tilde {\mathbf{x}}$}
			\STATE {\small {\textbf{Procedure:} {Enhancer}}$ (\mathbf{x}) $
				}
			\begin{ALC@g}
					\STATE  \small{Initialize parameters $\mathbf{\Theta}$ of INN-RSIC with Xavier.}
					\FOR{$\text{epoch} \gets 1$ \textbf{to} $\text{num\_epochs}$ }
					\STATE	\small{\textcolor{blue}{/\//\ Compute forward loss:}} \\
					$ \small {L_{\text{dec}} = ||f_{\mathbf{\Theta}}(\mathbf{x})-\tilde {\mathbf{x}}||^2, }$ \\
					$ \small{L_{\text{case}} = -\log\left(\frac{1}{(2\pi)^{M/2}} \exp\left(-\frac{1}{2} ||\mathbf{z}||^2\right)\right), }$
					\STATE	\small{\textcolor{blue}{/\//\ Compute backward loss:}} \\
					$	\small{L_{\text{en}} = || f_{\mathbf{\Theta}}^{-1}(\operatorname{QM}(f_{\mathbf{\Theta}}(\mathbf{x})), \vec{\mathbf{z}}) - \mathbf{x} ||^2,} $ \\
					$  	\small {L_{\text{per}} = ||\operatorname{Vgg}(\hat{\mathbf{x}}) - \operatorname{Vgg}(\mathbf{x})||^2,} $
					\STATE	\small{\textcolor{blue}{/\//\ Compute total loss:}} \\
					$	\small {L_{\text{total}} = \lambda_1 L_{\text{dec}} + \lambda_2 L_{\text{case}} + \lambda_3 L_{\text{en}} + \lambda_4 L_{\text{per}},} $   
					\STATE \small{\textcolor{blue}{/\//\  Update $\mathbf{\Theta}$ using gradient descent:}} \\
					$\small {\mathbf{\Theta} \gets \mathbf{\Theta} - \alpha \nabla L_{\text{total}}(\mathbf{\Theta})},$	
					\ENDFOR	
				\end{ALC@g}
			\STATE {\small {\textbf{Return}}: $\mathbf{\Theta}$}	 
		\end{algorithmic}
	\vspace{-0.5\baselineskip} % 调整垂直间距
	\hrulefill   \\ 
	\hspace*{0.02in}{\bf Inference Stage:}\\
	\hspace*{0.2in}\small{Input:}
	$ \mathbf{x}, \lambda $ \\
	\hspace*{0.2in}\small{Output:} 
	$\tilde {\mathbf{x}}, \hat {\mathbf{x}}$ 
	\begin{algorithmic}[1] 
			\STATE  {\small {\textbf{Procedure:}  Compressor}$ (\mathbf{x}, \lambda)$}
			\begin{ALC@g}		
					\STATE \small{$\tilde {\mathbf{x}} = \operatorname{ELIC}(\mathbf{x}, \lambda),  $}
				\end{ALC@g}	
			\STATE {\small {\textbf{Return}}: $\tilde {\mathbf{x}}$}
			\STATE {\small {\textbf{Procedure:} {Enhancer}}$ (\tilde {\mathbf{x}}) $}
			\begin{ALC@g}
					\STATE  \small{Loading the trained $\mathbf{\Theta}$ for INN-RSIC.} \\
					\STATE	\small{\textcolor{blue}{/\//\ Randomly sample $\vec{\mathbf{z}}$ and conduct inverse mapping:}} \\
					$ \vec {\mathbf{z}} \sim \mathcal{N}(0, \mathbf{I}),$ \\
					$\hat {\mathbf{x}} = f_{\mathbf{\Theta}}^{-1}(\tilde {\mathbf{x}}, \vec{\mathbf{z}}),$ \\	
				\end{ALC@g}	 
			\STATE {\small {\textbf{Return}: $\hat {\mathbf{x}}$}}
		\end{algorithmic} 
\end{algorithm}
\section{Experiments}\label{Sec:Experiments}
\subsection{Experimental Settings}\label{Sec:Experimental_Settings}
\subsubsection{Datasets} In this experiment, two datasets, including DOTA \cite{Ding_2019_CVPR} and UC-Merced (UC-M) \cite{yang2010bag}, are used for performance evaluation. Concretely, we use the training dataset of DOTA and 80\% of the UC-M training set for model training. Each image is randomly cropped into the resolution of $128 \times 128$. Random horizontal flip, random vertical flip, and random crop are applied for data augmentation. We randomly choose 100 images and 10\% images from the DOTA testing dataset and UC-M, respectively. Then each image is centrally cropped into  the resolution of $256 \times 256$ as the testing set.

\subsubsection{Implementation Details} In our experiment, we utilize ELIC \cite{he2022elic} to compress the images of the training dataset. Concretely, five ELIC models with parameters $\lambda \in \{4, 8, 32, 100, 450\} \times 10^{-4}$ are used to separately derive the decoded images, which are then utilized as labeled images to constrain the forward latent representation. 
As different compression rates result in different compression distortion distributions of ELIC, the labeled images under each $\lambda$ are used to train the INN-RSIC model. Additionally, to train the proposed model, the widely-used AdamW \cite{loshchilov2018decoupled} with $\beta_1$ = 0.9 and $\beta_2$ = 0.999 is employed for parameter optimization. 
The hyperparameters $\lambda_1$, $\lambda_2$, $\lambda_3$, and $\lambda_4$ are experimentally set to $1.0$, $200.0$, $0.05$, and $0.01$, respectively. We set the initial learning rate $\alpha$ to 1e-4 and the total number of epochs to 300, with the learning rate halved every 60 epochs until it reaches a value smaller than 1e-6. PSNR, MS-SSIM, FID \cite{heusel2017gans}, and LPIPS are adopted as the evaluation metrics.
\begin{figure*}[!htbp]
	\centering
	\subfigure[\fontsize{6}{10}\selectfont Evaluation on DOTA]{
		\begin{minipage}[t]{0.242\textwidth}
			\centering
			\includegraphics[scale=0.6]{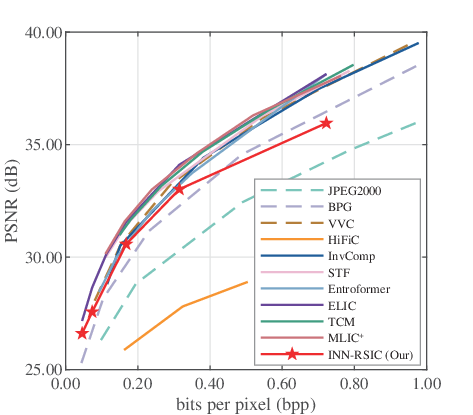}			
		\end{minipage}
		\begin{minipage}[t]{0.242\textwidth}
			\centering
			\includegraphics[scale=0.6]{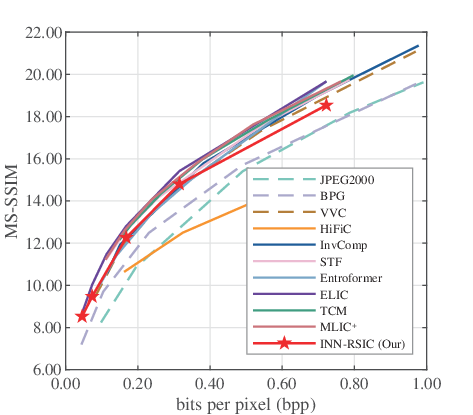}
		\end{minipage}
		\begin{minipage}[t]{0.242\textwidth}
			\centering
			\includegraphics[scale=0.6]{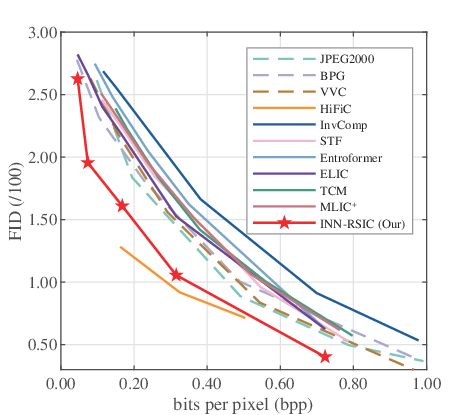}			
		\end{minipage}
		\begin{minipage}[t]{0.242\textwidth}
			\centering
			\includegraphics[scale=0.6]{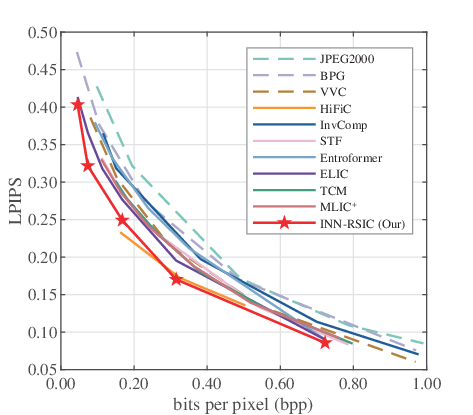}
		\end{minipage}
	}%
	\vspace{-2.2mm}
	\subfigure[\fontsize{6}{10}\selectfont Evaluation on UC-M]{
		\begin{minipage}[t]{0.242\textwidth}
			\centering
			\includegraphics[scale=0.6]{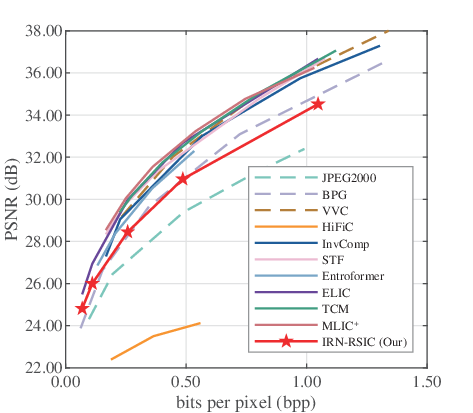}			
			%\caption{fig2}
		\end{minipage}
		\begin{minipage}[t]{0.242\textwidth}
			\centering
			\includegraphics[scale=0.6]{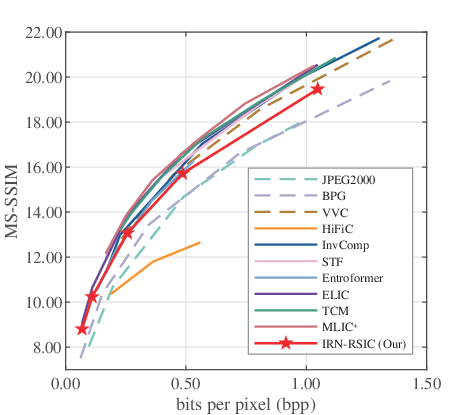}		
			%\caption{fig2}
		\end{minipage}
		\begin{minipage}[t]{0.242\textwidth}
			\centering
			\includegraphics[scale=0.6]{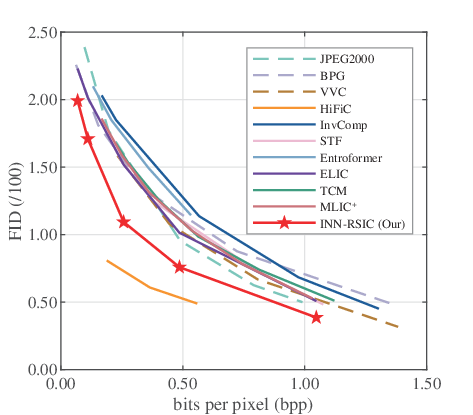}			
			%\caption{fig2}
		\end{minipage}
		\begin{minipage}[t]{0.242\textwidth}
			\centering
			\includegraphics[scale=0.6]{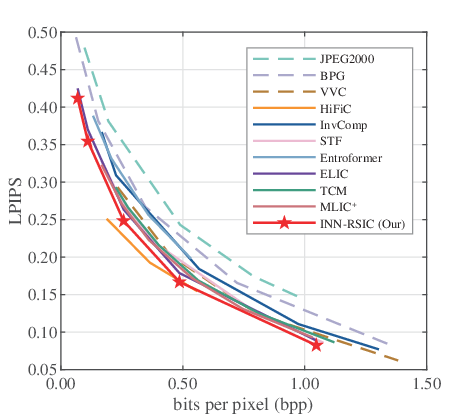}		
			%\caption{fig2}
		\end{minipage}
	}%
	\centering
	\caption{Performance comparison on the testing set of DOTA and UC-M. It is observed that INN-RSIC achieves a better trade-off between fidelity (\textit{i.e.}, higher PSNR) and perceptual quality compared to HiFiC.}
	\label{Fig:RD_DOTA}
\end{figure*}
\subsection{Performance Evaluation}
In this section, we conduct a comprehensive comparison of the proposed INN-RSIC and several competitive image compression algorithms, using the testing set of DOTA and UC-M. The benchmark includes the state-of-the-art image compression algorithms, including traditional standards such as JPEG2000 \cite{taubman2002jpeg2000}, BPG \cite{bpg2017}, VVC (YUV 444) \cite{VVC2021}, as well as the competitive learning-based image compression algorithms like HiFiC \cite{mentzer2020high}, InvComp \cite{xie2021enhanced},  STF \cite{zou2022devil},  Entrorformer \cite{qian2022entroformer}, ELIC \cite{he2022elic}, TCM \cite{liu2023learned}, and MLIC$ ^+ $ \cite{jiang2023mlic}. \par

The performance comparisons of these algorithms on the testing set of DOTA and UC-M are shown in Fig. \ref{Fig:RD_DOTA}. The results indicate that the proposed INN-RSIC outperforms all other methods in terms of FID and LPIPS, with the exception of HiFiC. Although HiFiC demonstrates comparable perceptual performance, it suffers from poor image fidelity, as evidenced by its low PSNR and MS-SSIM values. Additionally, at very low bpp, the improvements in both FID and LPIPS are minimal. This may be attributed to the high distortion of the decoded image, which makes it challenging for the proposed INN-RSIC to reliably recover more details from the Gaussian distribution conditioned on the decoded image. As the bpp increases, the proposed INN-RSIC shows significantly superior performance in terms of FID and LPIPS.
\par

\begin{figure*}[!htbp]
	\centering
	\subfigure[\fontsize{8}{10}\selectfont Testing images ``P0121" and ``P0216" of DOTA]{
		\begin{minipage}[t]{1\textwidth}
			\centering
			\includegraphics[scale=1.2]{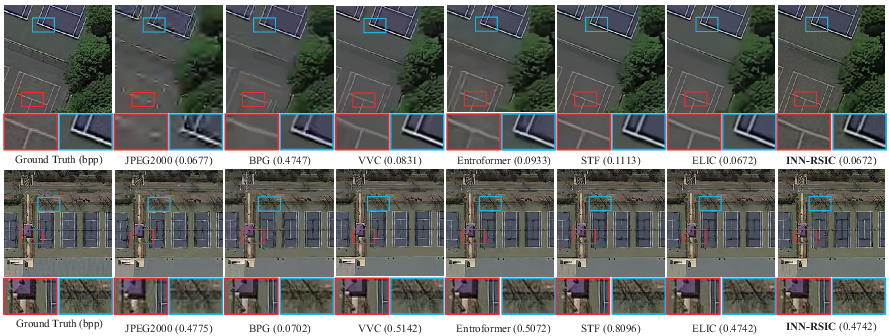}
			%			\caption{}
		\end{minipage}
	}%
	\vspace{-2.2mm}
	\subfigure[\fontsize{8}{10}\selectfont Testing images ``baseballdiamond95" and ``tenniscourt90" of UC-M]{
		\begin{minipage}[t]{1\textwidth}
			\centering
			\includegraphics[scale=1.2]{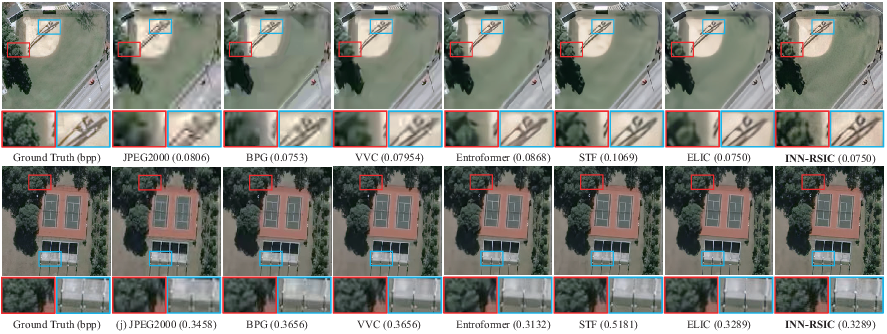}
			%\caption{fig1}
		\end{minipage}%
	}%
	\centering		
	\caption{Visualization of the decoding performance on the testing images of DOTA and UC-M.}
	\label{Fig:vis_com_DOTA}
\end{figure*}
Additionally, Fig. \ref{Fig:vis_com_DOTA} shows the decoding results of these algorithms on the testing images of DOTA and UC-M at low and high bitrates. In Fig. \ref{Fig:vis_com_DOTA}(a), it's noticeable that the image ``P0121" decoded by the state-of-the-art traditional compression algorithm VVC exhibits obvious distortion, even with an additional 23.66\% increase of bpp compared to the proposed INN-RSIC. Moreover, the decoded image from the competitive learning-based compression algorithm STF still suffers from significant distortion, despite a 70.73\% increase of bpp. At high bitrates, the decoded image ``P0216" by VVC presents blurry texture details despite the higher bpp. Particularly, it is evident that the perception quality of the decoded images by STF remains inferior to that of the proposed INN-RSIC, even with a 65.63\% increase in additional bpp. Similarly, as depicted in Fig. \ref{Fig:vis_com_DOTA}(b), the results further show that the proposed INN-RSIC demonstrates impressive perception quality compared to traditional and deep learning-based image compression algorithms at both low and high bitrates. It should be noted that, as shown in Fig. \ref{Fig:RD_DOTA}, since the compression performances of ELIC, TCM, and MLIC$ ^+ $ are comparable, for simplicity, we do not visualize the results of TCM and MLIC$ ^+ $.

In summary, we can safely demonstrate that the proposed INN-RSIC effectively and impressively contributes to the detailed recovery of decoded images without increasing additional bitrates.

\begin{table}[htbp]
	\renewcommand\arraystretch{1.0}
	\centering
	\caption{Quantitative comparison of the proposed INN-RSIC with and without QM or CGM}
	%	\resizebox{0.59\textwidth}{!}{%
		\begin{tabular}{cccccc}
			\toprule
			Datasets & Methods  & FID $\downarrow$ & LPIPS $\downarrow$ \\
			\hline		
			\multirow{3}{*}{DOTA} &  w/o QM & 236.8280 & 0.3521 \\
			&  w/o CGM & 235.6634 & 0.3496 \\
			& INN-RSIC & 195.6175 & 0.3218 \\
			\multirow{3}{*}{UC-M} &  w/o QM & 174.5280 & 0.3600 \\
			&  w/o CGM & 173.9155 & 0.3556 \\
			& INN-RSIC &  170.9994 & 0.3547 \\
			\bottomrule
		\end{tabular}%
		%	}
	\label{tab:aba_QM}%
\end{table}%
\subsection{Ablation Evaluation}
%In this section, we conduct ablation studies to assess the effectiveness of QM and CGM within the proposed INN-RSIC.
\subsubsection{Effectiveness of QM} QM considers the impact of format mismatch on the reconstruction performance of the inverse mapping.
Table \ref{tab:aba_QM} reports the quality metrics of the proposed model with and without QM for enhancing the decoded images of ELIC on the testing set of DOTA and UC-M. From the results, it can be seen that the proposed model with QM presents better perception quality results with comparable PSNR and MS-SSIM metrics on both datasets. The reason lies in the fact that the input images fed into the proposed INN-RSIC are 3-channel 8-bit RGB images during the inference stage, and QM can better contribute to the proposed model to match the data type of the input images. Moreover, the lower PSNR and MS-SSIM values of the enhanced images originate from the inherent reversibility of INN. Since the ground truth is not available at the decoding end, \textit{i.e.}, the synthetic image is not available, we therefore use the decoded image as the inverse input to the proposed INN-RSIC in the inference stage. 

%The difference between the synthetic and decoded images makes it difficult to obtain high PSNR and MS-SSIM results.
%Additionally, Table \ref{tab:aba_CGM} reports the quality metrics of the proposed model with and without CGM on the testing set of UC and DOTA datasets. From the results, it can be observed that the proposed INN-RSIC with CGM presents better perception quality results in terms of FID and LPIPS with comparable PSNR and MS-SSIM metrics on both datasets.
\subsubsection{Effectiveness of CGM} CGM aims to incorporate guidance from synthetic images into the latent variables during the reconstruction of texture-rich images. To demonstrate the effectiveness of this approach. Table \ref{tab:aba_QM} presents the performance of the proposed INN-RSIC with and without CGM on the testing sets of DOTA and UC-M. It is evident that the INN-RSIC with CGM consistently outperforms the model without CGM in terms of FID and LPIPS. This implies that encoding latent variables into Gaussian distributed variables conditioned on synthetic images can enhance the reconstruction performance during inverse mapping. This observation is also consistent with the earlier research  \cite{9467808}. \par

\begin{figure}[!htbp]
	\centering
	\includegraphics[scale=0.58]{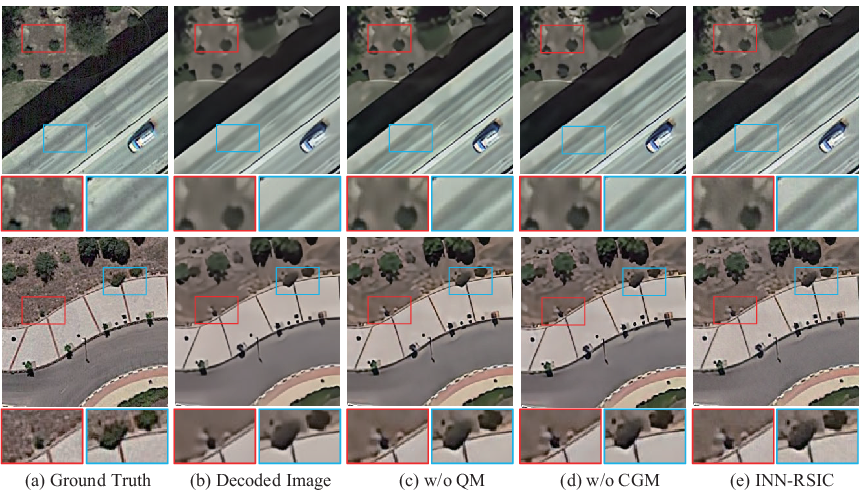}
	\caption{Visualization of the enhanced images generated by the proposed INN-RSIC with and without (w/o) QM or CGM on the testing images ``P0088" and ``P0097" of DOTA. ELIC is adopted to derive the decoded images, which serve as input for the proposed INN-RSIC for enhancement.}
	\label{Fig:abla_QM}
\end{figure}

To visualize the effectiveness of QM and CGM, Fig. \ref{Fig:abla_QM} shows the enhanced images of the proposed INN-RSIC with and without QM or CGM on the decoded images ``P0088" and ``P0097" by ELIC. The results indicate that although the enhanced images have higher perceptual quality than the decoded images when either QM or CGM is not available, the proposed INN-RSIC provides the best perceptual quality when QM and CGM are used.

\subsection{Robustness Evaluation}\label{Sec:Robustness}
\begin{figure}[!htbp]
	\centering
	\subfigure[\fontsize{8}{10}\selectfont Robust performance trained at low-bitrate distortion]{
		\begin{minipage}[t]{0.245 \textwidth}
			\centering
			\includegraphics[scale=0.57]{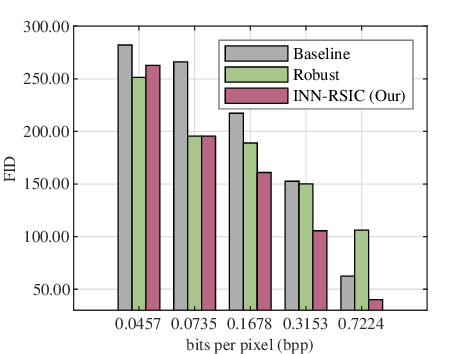}
			%\caption{fig1}
		\end{minipage}%
		
		\begin{minipage}[t]{0.245 \textwidth}
			\centering
			\includegraphics[scale=0.57]{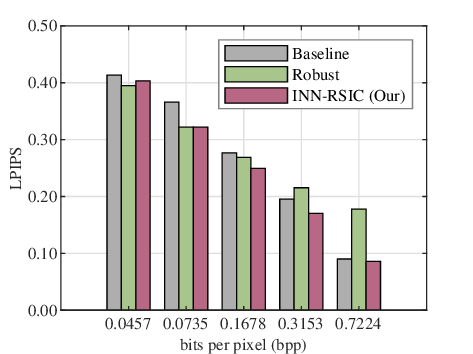}
			%\caption{fig2}
		\end{minipage}
	}%	
		\vspace{-1.6mm}
	\subfigure[\fontsize{8}{10}\selectfont Robust performance trained at high-bitrate distortion]{
		\begin{minipage}[t]{0.245 \textwidth}
			\centering
			\includegraphics[scale=0.57]{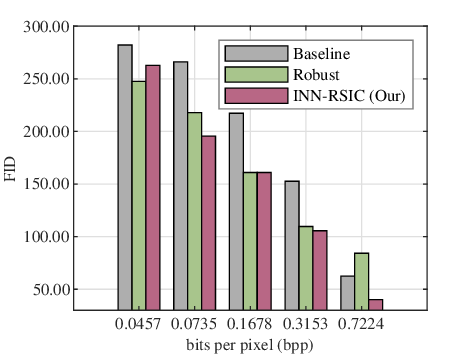}
			%\caption{fig1}
		\end{minipage}%
		\begin{minipage}[t]{0.245 \textwidth}
			\centering
			\includegraphics[scale=0.57]{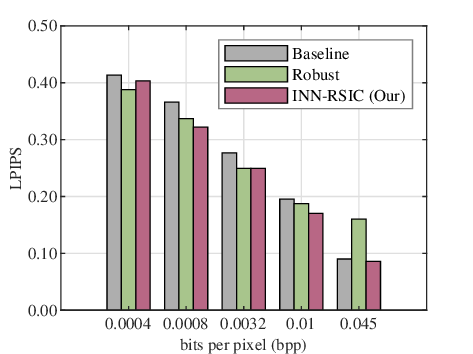}
			%\caption{fig2}
		\end{minipage}
	}%	
	\centering
	\caption{FID and LPIPS comparison of the baseline, robust, and our INN-RSIC. The baseline refers to testing on the decoded images of ELIC without additional improvement. The robust case indicates that the proposed INN-RSIC trained at bpp 0.0735 (a) or bpp 0.1678 (b) is adopted to enhance the decoded images at a wide range of bpp.}
	\label{Fig:0.0008_for_robust}
\end{figure}

The robustness of a model plays a critical role in practical deployment. Here, we employ two INN-RSIC models trained at two values of $\lambda$ to enhance the decoded images of ELIC across various compression ratios. The proposed INN-RSIC trained at $\lambda = 8 \times 10^{-4} $ is used to enhance the images compressed and decoded using ELIC at different $\lambda$ for the images in the testing set of DOTA, where $\lambda \in \{4, 8, 32, 100, 450\} \times 10^{-4} $. FID and LPIPS comparison results are depicted in Fig. \ref{Fig:0.0008_for_robust}(a). Comparing the FID values between the baseline and robust cases, it is observed that except for $\lambda =450 \times 10^{-4}$, the proposed model trained at $\lambda = 8 \times 10^{-4}$ can still improve the perception quality of others.\par
%This suggests the robustness of the proposed INN-RSIC in enhancing the perception quality of decoded images. 

Besides, the proposed INN-RSIC trained with the corresponding $\lambda$ demonstrates the most significant improvement in perception quality in terms of FID and LPIPS, except for $\lambda = 4 \times 10^{-4}$.  The poorer performance of the proposed INN-RSIC at low bpp $0.0457$ (\textit{i.e.}, $\lambda=4 \times 10^{-4}$) may be because the quality of the corresponding decoded image is poor at very low bpp, whereas the Gaussian distributed variables are conditioned on the decoded image. That is, the decoded image has limited guidance to recover stable results. \par
\begin{figure}[!htbp]
	\centering
	\subfigure[\fontsize{6}{10}\selectfont Testing images ``P0195" and ``P0059"]{
		\begin{minipage}[t]{0.5\textwidth}
			\centering
			\includegraphics[scale=1.15]{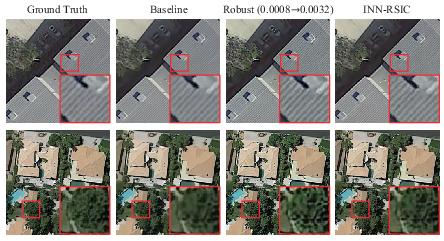}
			%\caption{fig1}
		\end{minipage}%
	}%	
	\vspace{-2mm}
	\subfigure[\fontsize{6}{10}\selectfont Testing images ``P0015" and ``P0089"]{
		\begin{minipage}[t]{0.5\textwidth}
			\centering
			\includegraphics[scale=1.15]{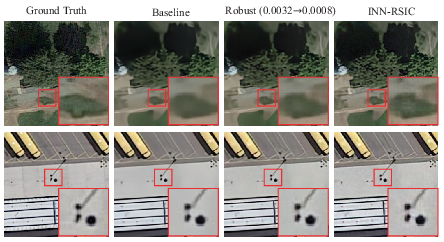}
			%\caption{fig2}
		\end{minipage}
	}%
	\centering
	\caption{Robust performance comparison on the testing images of DOTA. (a) Low-bitrate ($\lambda = 8 \times 10^{-4}$) trained model to enhance the decoded images at high bitrate ($\lambda = 32 \times 10^{-4}$); (b) High-bitrate trained model ($\lambda = 32 \times 10^{-4}$) to enhance the decoded images at low bitrate ($\lambda = 8 \times 10^{-4}$).}
	\label{Fig:Robust_0.0008_0.0032}
\end{figure}
Furthermore, we investigate the robustness of the proposed INN-RSIC model trained at high bitrate conditions in improving decoded images across different bitrate levels. The proposed INN-RSIC trained at $\lambda = 32 \times 10^{-4}$ is used to enhance the images compressed and decoded using ELIC at different $\lambda$ for the images in the testing set of DOTA, where $\lambda \in \{4, 8, 32, 100, 450\} \times 10^{-4} $. FID and LPIPS comparison results are illustrated in Fig. \ref{Fig:0.0008_for_robust}(b). Comparing the FID values between the baseline and robust cases, it is observed that except for $\lambda = 450 \times 10^{-4}$, the proposed model trained at $\lambda = 32 \times 10^{-4}$ can still improve perception quality across different bitrate levels. Moreover, the proposed INN-RSIC trained with the corresponding $\lambda$ demonstrates the most significant improvement in perception quality in terms of FID and LPIPS, except for $ \lambda = 4 \times 10^{-4}$.\par

To visualize the robust performance of the model, the robust case in Figs. \ref{Fig:Robust_0.0008_0.0032} (a) and (b) show the enhanced images obtained by augmenting the decoded images of ELIC at high and low bitrates using the INN-RSIC model trained at low and high bitrates, respectively. The baseline denotes the image decoded by ELIC. From the results, it can be observed that although the decoded image of the robust case presents more artifacts and burrs compared to the ground truth and the proposed INN-RSIC, it presents better visual characteristics and recovers more texture information compared to the baseline on the global vision.

In summary, the results indicate that both the proposed models trained at low and high bitrates exhibit effective enhancement of perception quality for decoded images across a wide range of bpp levels, highlighting the impressive robustness of the proposed INN-RSIC in enhancing decoded images.
\begin{figure*}[!htbp]
	\centering
	\includegraphics[scale=1.12]{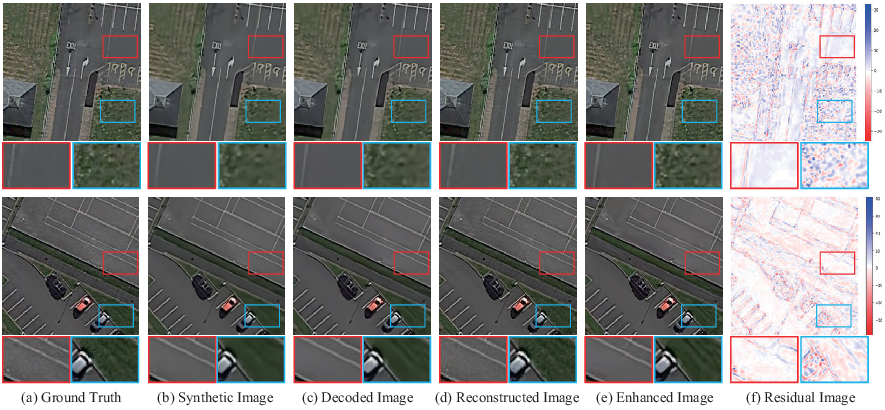}
	\caption{Visualization of images generated by the proposed INN-RSIC with different inputs during the inference stage under $\lambda=32\times10^{-4}$. The reconstructed images are obtained using the synthetic images as input, while the enhanced images are derived using the decoded images as input. The residual images illustrate the difference between the synthetic images and the decoded images.}
	\label{Fig:input_Sensitivity}
\end{figure*}
\subsection{Discussion}\label{Sec:sensitivity}
During training, we utilize decoded images as labels to guide the forward training of INN-RSIC in generating synthetic images, which are then fed into the inverse network along with Gaussian-distributed latent variables for image reconstruction. However, in the inference stage, as the ground truth cannot be obtained to feed into the forward network at the decoding end, \textit{i.e.}, the synthetic image cannot be derived, so we resort to using labeled images instead of synthetic ones. In other words, the decoded images are fed into the inverse network along with Gaussian-distributed latent variables for enhancing the texture of the decoded images. While the texture information of the enhanced images is enriched with the assistance of Gaussian-distributed latent variables, the strictly invertible characteristic of INN inevitably affects the reconstruction quality of the enhanced image when using decoded images instead of synthetic ones. \par

This investigation primarily focuses on exploring the input sensitivity of the proposed INN-RSIC model. Fig. \ref{Fig:input_Sensitivity} showcases the images restored by INN-RSIC with synthetic images and decoded images as inputs during the inference stage. The results reveal that when synthetic images are used as input, the reconstructed image effectively recovers texture-rich details. However, when decoded images are employed as input, although some details are still recovered in the enhanced image, noticeable detail distortions emerge, particularly in texture-rich regions. To explore this issue further, we analyze the residual image between the synthetic image and the decoded image. It becomes evident that the differences contradict the invertible characteristic of INN, thereby inevitably limiting the performance of the final texture recovery. \par

In summary, the similarity between decoded and synthetic images significantly influences the recovery of texture-rich image details. Optimizing the algorithm design to ensure that the resulting synthetic and decoded images are as similar as possible will greatly enhance the decoding of detailed texture-rich RS images without requiring additional bitrates. This optimization holds the potential to improve the overall performance and fidelity of the INN-RSIC model, allowing for a more accurate reconstruction of RS images.

\subsection{Model Complexity}\label{Sec:Complexity}
To assess the computational complexity of the proposed method, we conduct the experiments on the DOTA testing set with an Intel Silver 4214R CPU running at 2.40GHz and one NVIDIA GeForce RTX 3090 Ti GPU. The average results are given in Table \ref{tab:parameter_FLOP}. \par
\begin{table*}[!htbp]
	\renewcommand\arraystretch{1.1}
	\centering
	\caption{Compare the parameters and computational cost among various image compression algorithms.}
	\begin{tabular}{ccccccc}
		\toprule
		\multirow{2}{*}{Methods} & \multirow{2}{*}{Parameters (M)} & \multirow{2}{*}{FLOPs (G)} & \multicolumn{2}{c}{CPU} & \multicolumn{2}{c}{GPU} \\
		\cline{4-7}        &     &     & Encoding Times (s) & Dec. Times (s) & Enc. Times (s) & Dec. Times (s) \\
		\hline
		HiFiC & 55.42 & 148.24 & 7.1366 & 10.1388 & 0.4065 & 0.8096 \\
		InvComp & 67.12 & 41.81 & 39.2739 & 42.0858 & 1.2498 & 3.6382 \\
		Entroformer & 12.67 & 44.76 & 1.3482 & 0.5018 & 0.2762 & 0.0919 \\
		STF & 33.35 & 99.83 & 1.1395 & 1.3175 & 0.1343 & 0.1879 \\
		ELIC & 54.46 & 31.66 & 1.7934 & 1.6796 & 0.4671	 & 0.2507 \\
		TCM & 44.97 & 35.23 & 22.0959 & 21.0951 & 0.4637 & 0.4045 \\
		MLIC$ ^+ $ & 82.36 & 116.48 & 38.5316 & 38.6344 & 0.3678 & 0.5007 \\
		IRN-RSIC & 55.71 & 63.57 & 1.7934 & 2.1616 & 0.4671 & 0.2729 \\
		\bottomrule
	\end{tabular}%
	\label{tab:parameter_FLOP}%
\end{table*}%
From the results, it is clear that MLIC$^+$ has 82.36M parameters, whereas the proposed INN-RSIC comprises only 67.64\% of that count. Notably, the INN-based enhancer we introduced adds only 1.25M parameters, representing a mere 2.30\% increase over the baseline ELIC, which underscores its remarkable memory efficiency. 
In terms of floating point operations per second (FLOPs), HiFiC and MLIC$^+$ bear a heavy computational load, with FLOP values of 148.24G and 116.48G, respectively. In contrast, the proposed INN-RSIC requires only 42.88\% and 54.58\% of the FLOPs of HiFiC and MLIC$^+$, respectively.
As for the decoding times, one can observe that the GPU decoding time of the proposed method is just 11.03\% and 8.91\% of that of the competitive TCM and MLIC$^+$ algorithms, respectively.

To sum up, the proposed method requires less memory space, has lower computational complexity, and enjoys outstanding decoding speed on both CPU and GPU.

\section{Conclusion} \label{Sec:Conclusion}
In this paper, we present INN-RSIC, a novel method designed to enhance the perceptual quality of decoded RS images. This approach encodes the compression distortion of an image compression algorithm into Gaussian-distributed latent variables and  leverages invertible transformations to substantially restore texture details. Additionally, we introduce the CCGM to separate and encode the compression distortion from the ground truth into latent variables conditioned on the synthetic image, thereby improving the reconstruction of the enhanced image. Moreover, we incorporate QM to mitigate the impact of data type conversions on reconstruction quality during inference. Extensive experiments on two widely used RS image datasets demonstrate that INN-RSIC achieves a superior balance between perceptual quality and high image fidelity compared to state-of-the-art image compression algorithms. Furthermore, our approach offers a new research perspective for enhancing the texture reconstruction capability of RS image compression algorithms by minimizing the gap between synthesized and decoded images.

%In the future, we will optimize the proposed INN-RSIC to minimize the gap between synthetic and decoded images, thereby improving the detail recovery performance of the proposed INN-RSIC for RS image compression.

%\newpage
%\appendix
%\section{Supplemental Material}
%In this supplementary material, Section \ref{Sec:Robustness} evaluates the robustness of the proposed INN-RSIC initially. Next, in Section \ref{Sec:sensitivity}, we analyze the limitations of INN-RSIC and propose a research direction to further improve the texture details of decoded RS images. Subsequently, in Section \ref{Sec:AM_CGM_vis}, we visualize the effectiveness of QM and CGM within the proposed INN-RSIC. Finally, Section \ref{Sec:Complexity} presents the model complexity.

\bibliographystyle{IEEEtran}
\bibliography{reference.bib}

\end{document}